\documentclass[conference]{IEEEtran}
\usepackage[pdftex]{graphicx}
\usepackage{svg}
\usepackage{titlesec}
\titlespacing*{\section}{0pt}{0.8ex}{0.4ex}
\usepackage{cite}
\usepackage{amsmath}
\usepackage{amsfonts}
\usepackage{amssymb}
\usepackage{booktabs}
\usepackage{subcaption}
\usepackage{multirow}
\usepackage{bm}
\usepackage{xcolor}
\usepackage{tikz}
\usepackage{adjustbox}
\usepackage{amsthm}
\newtheorem{remark}{Remark}

\usepackage[table]{xcolor}
\IEEEoverridecommandlockouts
\usetikzlibrary{shapes.geometric, arrows.meta, positioning, calc}
\usepackage{url}
\usepackage[dvipsnames]{xcolor} 
\usepackage{hyperref}
\hypersetup{
    colorlinks=true,
    linkcolor=magenta, 
    urlcolor=magenta    
}

\begin{document}

\title{Physics-Guided Self-Supervised Statistical Residual Learning for Sonar Despeckling with Improved Generalization}

\author{
    Swapna~Pillai,
    Siddharth~Singh~Savner,
    Sujit~Kumar~Sahoo%
    \thanks{S. Pillai and S. K. Sahoo are with the School of Electrical Sciences, Indian Institute of Technology Goa, Goa 403401, India (e-mail: swapna23242104@iitgoa.ac.in; sujit@iitgoa.ac.in). }%
    \thanks{S. S. Savner is with Inria, Sophia Antipolis, France (e-mail: siddharth-singh.savner@inria.fr). }%
    \thanks{The code will be made available at: \url{https://github.com/SwapnaPillai55/Despeckling_SSL}} 
}

\maketitle

\begin{center}
\fbox{
\parbox{0.95\linewidth}{
\small
\copyright~2026 IEEE. Personal use of this material is permitted.
Permission from IEEE must be obtained for all other uses, in any
current or future media, including reprinting/republishing this
material for advertising or promotional purposes, creating new
collective works, for resale or redistribution to servers or lists,
or reuse of any copyrighted component of this work in other works.
}
}
\end{center}

\begin{abstract}
This letter introduces a physics-informed self-supervised framework for sonar image despeckling that reformulates despeckling as residual consistency in the homomorphic log domain. By constraining the log-ratio residual to obey multiplicative speckle statistics, the proposed method eliminates the need for clean supervision while preventing degenerate identity solutions. A variance-targeted statistical loss combined with edge-aware structural regularization and median-guided curriculum stabilization enables effective speckle suppression with preserved structural fidelity. This formulation along with a lightweight neural network achieves state-of-the-art performance across multiple real sonar datasets and demonstrates excellent cross-dataset robustness, while remaining suitable for real-time deployment.
\end{abstract}


\section{Introduction}
\IEEEPARstart{S}{peckle} noise is intrinsic to coherent imaging systems such as Synthetic Aperture Radar (SAR) and Synthetic Aperture Sonar (SAS), arising from interference within a resolution cell \cite{Speckle}. It follows a multiplicative model and induces strong local fluctuations that reduce interpretability and degrade downstream detection/classification performance \cite{SP_Oceans24}. Despeckling remains fundamentally challenging due to the absence of physically realizable speckle-free ground truth.

Traditional despeckling filters rely on local statistics to trade noise suppression against edge preservation . Multiresolution models improve scale adaptivity \cite{Philippe_wavelet, argenti2006multiresolution,achim2006sar}, and non-local methods exploit self-similarity at higher computational cost \cite{PPB,BM3D_2}. Although effective in controlled settings, these hand-crafted approaches often require careful tuning and can be brittle across acquisition conditions. Recent deep learning methods offer strong performance but introduce new limitations. Supervised models trained on synthetic speckle pairs \cite{IDCNN,SAR_CNN, mestre2025deep} fail to generalize to real sonar due to mismatched noise statistics and unknown preprocessing. Self-supervised methods avoid clean targets \cite{noise2noise,N2V,S2V, GSSM,SEGSID}, yet many depend on blind-spot masking or independence assumptions that are inconsistent with spatially correlated speckle, leading to over-smoothing or degenerate identity solutions.

In this letter, we propose a physics-informed self-supervised framework that formulates sonar despeckling as residual consistency in the homomorphic log domain. Rather than regressing a clean image, the network learns residuals constrained to satisfy physically consistent multiplicative speckle statistics under explicit variance-targeted regularization, preventing degenerate identity solutions. An edge-aware structural term with curriculum stabilization preserves fine-scale details during optimization. The resulting compact architecture achieves state-of-the-art M-score across multiple real sonar datasets without requiring clean supervision.


\section{Signal Model and Statistical Background}
\label{sec:stat}

In coherent imaging, the observed intensity $y$ is modeled as the product of reflectivity $x$ and multiplicative speckle $n$ \cite{ulaby2019handbook}:
\begin{equation}
y = x \cdot n.
\end{equation}
For single-look systems $n$ follows an exponential distribution, while multi-look observations follow a Gamma distribution with unit mean. Applying a logarithmic transform yields the additive form
\begin{equation}
z = \ln(y) = \ln(x) + \eta, \quad \eta = \ln(n).
\end{equation}

The log-speckle variance is governed by the trigamma function $\psi(1,L)$, where $L$ denotes the equivalent number of looks \cite{trigamma}:
\begin{equation}
\label{tar_var}
\sigma_{\text{tgt}}^2 = \psi(1,L).
\end{equation}

In practice, despeckling quality is evaluated using locally normalized residual statistics such as the M-score \cite{Gomez2017}, where homogeneous regions are automatically selected using local mean and ENL consistency criteria on the ratio image. The metric measures residual variance consistency within these regions, motivating the use of variance-targeted residual constraints during training to align the learning objective with physically meaningful speckle statistics and established evaluation criteria.
\section{Methodology}
\label{sec:method}

\begin{figure*}[t]
    \centering
    \resizebox{1.1\textwidth}{!}{%
        \includegraphics [trim={2cm 10.5cm 0 0.5cm},
            clip
        ]{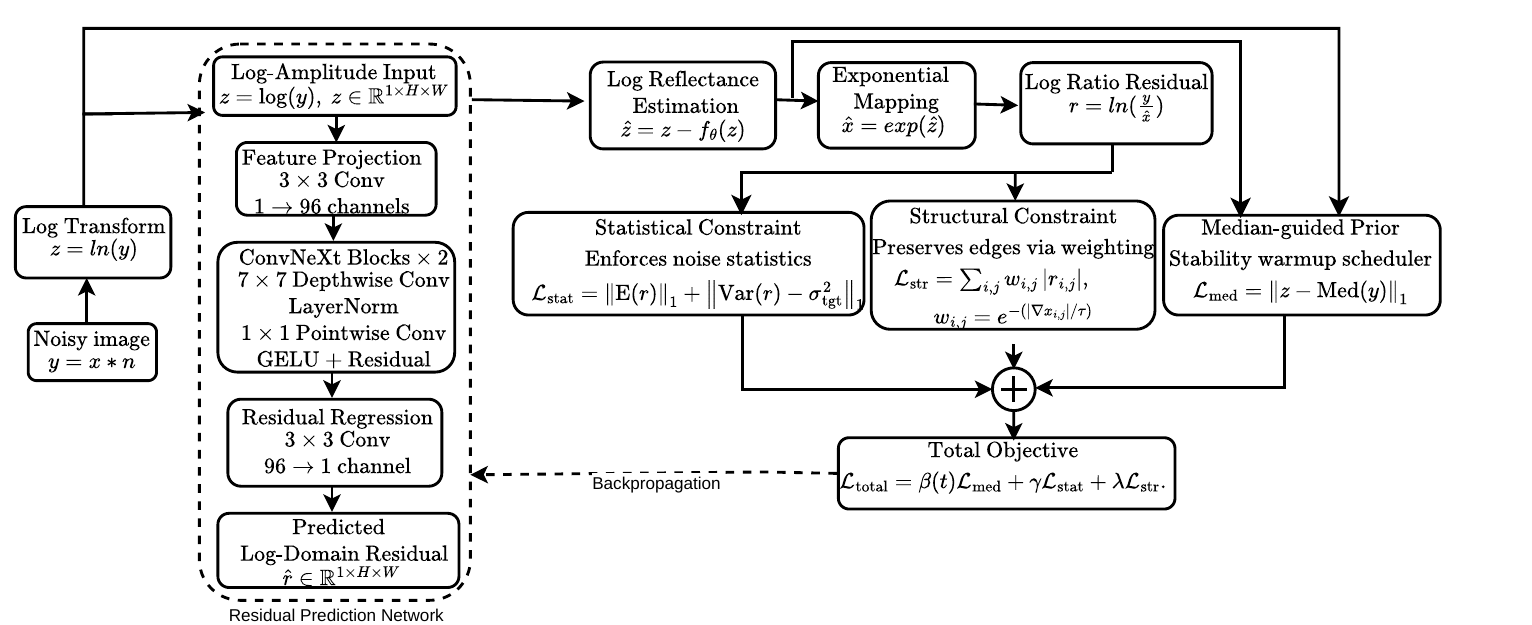}
        
    }
    \caption{Overview of the proposed self-supervised despeckling architecture. The residual prediction network takes a log-amplitude noisy image as input and estimates an additive residual $\hat{r}$ in the log-domain, consistent with the multiplicative speckle noise model. 
$\mathcal{L}_{\mathrm{stat}}$, $\mathcal{L}_{\mathrm{str}}$, and $\mathcal{L}_{\mathrm{med}}$ denote the statistical, structural, and curriculum losses, respectively; $\mathrm{Med}(y)$ is a spatial median-filtered observation and $\sigma_{\text{tgt}}^2$ is the target log-speckle variance.
$\beta$, $\gamma$, and $\lambda$ weight the corresponding loss terms.}
    \label{BD}
\end{figure*}


The proposed framework performs sonar despeckling via physics-guided self-supervised
residual learning in the homomorphic log domain (Fig.~\ref{BD}).
Given a noisy intensity image $y$, we apply the logarithmic transform
\begin{equation}
z=\ln(y),
\end{equation}
which converts multiplicative speckle into an additive component.
A residual prediction network (RPN) $f_\theta$ estimates the log-domain speckle
residual $f_\theta(z)$, and the clean log-reflectivity $\hat{z}$ is recovered as
\begin{equation}
\hat{z}=z-f_\theta(z).
\end{equation}
The resulting log-ratio residual
\begin{equation}
r=\ln\!\left(\frac{y}{\hat{x}}\right),
\end{equation}
ideally contains only stochastic speckle and forms the basis for self-supervised
statistical regularization, where $\hat{x}$ denotes the estimated despeckled image.

\subsection{Physics-Informed Statistical Constraint}

To prevent the degenerate identity solution ($\hat{x}=y$), we constrain the first- and second-order statistics of $r$:
\begin{equation}
\mathcal{L}_{\mathrm{stat}} = 
\|\mathbb{E}[r]\|_1 +
\|\mathrm{Var}(r) - \sigma_{\text{tgt}}^2\|_1 .
\end{equation}
The target variance $\sigma_{\text{tgt}}^2$ is empirically selected per dataset and closely matches theoretical log-speckle variance corresponding to its equivalent number of looks. This enforces physically consistent residual behavior while suppressing identity and over-smoothing solutions.

\subsection{Edge-Aware Structural Regularization}

To preserve fine structures, we introduce an edge-aware edge-weighted residual smoothness term:
\begin{equation}
\mathcal{L}_{\mathrm{str}} =
\sum_{i,j}
\exp\!\left(-\frac{\|\nabla \hat{x}_{i,j}\|}{\sigma}\right)
|\nabla r_{i,j}|.
\end{equation}
The exponential weight attenuates smoothing near strong gradients (salient targets) and activates it in homogeneous regions, ensuring speckle suppression without edge erosion.

\subsection{Curriculum-Stabilized Training}

Early training is stabilized using a median-guided prior
\begin{equation}
\mathcal{L}_{\mathrm{med}} =
\|\hat{z} - \ln(\mathrm{Med}(y)+\epsilon)\|_1,
\end{equation}
where the weight $\beta(t)$ decreases linearly over $T$ epochs. 

The final objective is:
\begin{equation}
\mathcal{L}_{\mathrm{total}} =
\beta(t)\mathcal{L}_{\mathrm{med}} +
\gamma \mathcal{L}_{\mathrm{stat}} +
\lambda \mathcal{L}_{\mathrm{str}}.
\end{equation}


\subsection{Theoretical Interpretation}

The proposed formulation admits a clear theoretical interpretation, linking residual alignment to variance suppression.

\begin{remark}[Speckle variance reduction in homogeneous regions]
Assume $\ln(x)$ is constant over a homogeneous region $\Omega$.
If training drives $\mathbb{E}[(r-\eta)^2]\to 0$ on $\Omega$,
then $\mathrm{Var}(\hat{z})\to 0$ on $\Omega$.
\end{remark}

\noindent\textit{Explanation.}
On $\Omega$, $\hat{z}=\ln(x)+(\eta-r)$, hence
$\mathrm{Var}(\hat{z})=\mathrm{Var}(\eta-r)$.
Mean-square convergence of $r$ to $\eta$ implies variance collapse.

In addition, the statistical constraint ensures identifiability of the solution.

\begin{remark}[Exclusion of the identity solution]
The identity mapping $\hat{x}=y$ cannot minimize
$\mathcal{L}_{\mathrm{stat}}$.
\end{remark}

\noindent\textit{Explanation.}
For $\hat{x}=y$, $r=0$ and $\mathrm{Var}(r)=0$, yielding
$\mathcal{L}_{\mathrm{stat}}=\sigma_{\mathrm{tgt}}^{2}>0$.
Any residual satisfying the target variance achieves lower loss.

Together, the statistical and structural terms act as complementary constraints:
the former enforces physical consistency, while the latter preserves spatial detail.

\section{Experiments}
\label{sec:experiments}

\subsection{Datasets}

Evaluation is conducted on four sonar datasets with diverse speckle characteristics. 
KLSG~\cite{HHUCzCz_SeabedObjectsKLSG} contains 447 side-scan sonar images (85 test) with relatively higher equivalent looks. 
DEBRIS~\cite{singh2021marine} (FLS data) comprises 1868 forward-looking sonar images (374 test) acquired using an ARIS Explorer 3000 system and exhibits stronger multiplicative fluctuations. 
URPC~\cite{URPC} includes 1197 underwater object images (192 test) with moderate speckle variance and complex scene structures. 
SASSED~\cite{SASSED} provides 129 synthetic aperture sonar image snippets characterized by Gamma-distributed speckle typical of multi-look coherent imaging.

\subsection{Residual Prediction Network (RPN)}

Given a noisy sonar image $y \in \mathbb{R}^{1\times H\times W}$, the despeckled image
$\hat{x}$ is obtained by operating in the log domain $z=\log(y)$ and estimating a
log-domain residual using a RPN denoted by $f_{\theta}$,
such that $\hat{x}=\exp\!\left(z-f_{\theta}(z)\right)$ (see Fig.~\ref{BD}).
$f_{\theta}$ follows an isotropic ConvNeXt-style architecture \cite{next}. In this, $3\times3$ projection first maps $z$ to 96 channels, followed by two residual ConvNeXt blocks:
\begin{equation}
\begin{aligned}
f_{k+1} = f_k + \gamma_k\,\mathcal{P}_{1\times1}\!\left(
\mathrm{GELU}\!\left(
\mathcal{P}_{1\times1}\!\left(
\mathrm{LN}\!\left(\mathcal{D}_{7\times7}(f_k)\right)
\right)\right)\right),
\end{aligned}
\end{equation}
where $k\in\{0,1\}$.
Here, $\mathcal{D}_{7\times7}$ and $\mathcal{P}_{1\times1}$ denote depth-wise and
point-wise convolutions, respectively. A final $3\times3$ layer produces the log-domain residual estimate. Predicting an additive residual in the homomorphic domain preserves the multiplicative speckle model while constraining representational capacity, thereby mitigating over-smoothing and suppressing identity mappings.


\subsection{Training Protocol}

The RPN is derived from ConvNeXt~\cite{next} and adapted for isotropic, full-resolution processing in the log-amplitude domain. The original patchifying downsampling stem is replaced with a stride-1 convolution to preserve spatial resolution for dense residual estimation. Training is performed on randomly extracted $64\times64$ patches with a batch size of 8. The isotropic design, based on uniform receptive fields, mitigates directional bias and ensures consistent treatment of sonar textures, which lack a preferred orientation. The network is optimized in the homomorphic domain to estimate the speckle residual rather than directly regress the clean image.

\section{Results and Discussion}
\label{sec:results}

Table~\ref{tab:mscore_sonar} presents a quantitative comparison of the proposed method with classical despeckling filters and recent self-supervised deep learning approaches on three representative sonar datasets using the M-score metric. The network is trained on the KLSG and DEBRIS datasets, while performance on URPC is evaluated using the model trained on DEBRIS. Despeckling performance is assessed using the M-score, which quantifies residual speckle variance consistency in homogeneous regions; lower values indicate more effective speckle suppression while preserving structural content. For the proposed statistical residual constraint, the target log-speckle variance $\sigma_{\text{tgt}}^2$ is set to $0.1175$ for KLSG (corresponding to $\text{NLooks}=9$), $0.0689$ for DEBRIS (corresponding to $\text{NLooks}=15$), and $0.0869$ for URPC (corresponding to $\text{NLooks}=12$), reflecting the dataset-specific speckle characteristics. These values were empirically estimated using M-score evaluation on a validation subset comprising $10\%$ of the training data, by evaluating candidate target variances corresponding to $\text{NLooks} \in [4,20]$ according to Eqn.~\ref{tar_var}.

\begin{table}[h]
\centering
\caption{Quantitative comparison of despeckling methods on sonar datasets using the M-score metric. For each dataset, the best (lowest) M-score is highlighted in bold, and the second best is underlined.}
\label{tab:mscore_sonar}
\setlength{\tabcolsep}{6pt}
\renewcommand{\arraystretch}{1.1}
\begin{tabular}{lccc}
\hline
\multirow{2}{*}{\textbf{Method}} & \multicolumn{3}{c}{\textbf{Datasets}} \\
\cline{2-4}
 & \textbf{KLSG} & \textbf{URPC} & \textbf{DEBRIS} \\
\hline
PPB~\cite{PPB} (TIP'09) & 9.683 & 7.824 & 6.265 \\
FANS~\cite{FANS} (GRSL'13) & 7.333 & 5.595 & 3.889 \\
Noise2Void~\cite{N2V} (CVPR'19) & 17.297 & 7.682 & 3.384 \\
Speckle2Void~\cite{S2V} (TGRS'21) & 30.638 & 22.038 & 20.887 \\
AP-BSN~\cite{Ap-bsn} (CVPR'22) & 8.879 & 6.407 & 4.187 \\
CVF-SID~\cite{Cvf-sid} (CVPR'22) & 7.969 & 5.704 & 3.138 \\
Blind2Unblind~\cite{blind2unblind} (CVPR'22) & 7.908 & 5.291 & 3.181 \\
GSSM~\cite{GSSM} (Neurocomputing'23) & 12.533 & 6.415 & 3.055 \\
SDAP~\cite{SDAP} (ICCV'23) & 8.703 & 7.018 & 4.988 \\
LG-BPN~\cite{LG-BPN} (CVPR'23) & 8.839 & 5.599 & 4.227 \\
PUCA~\cite{puca} (NeurIPS'23) & 7.745 & 6.702 & 4.889 \\
SS-BSN~\cite{SS-BSN} (IJCAI'23) & 7.662 & 6.605 & 4.156 \\
SEGSID~\cite{SEGSID} (TIP'25) & \underline{6.482} & 4.870 & 2.875 \\
SEGSID-KD~\cite{SEGSID} (TIP'25) & 6.492 & \underline{4.869} & \underline{2.769} \\ \hline
\textbf{Proposed} & \textbf{6.142} & \textbf{3.847} & \textbf{1.321} \\
\hline
\end{tabular}
\end{table}

To further evaluate structural fidelity, the Edge Preservation Index (EPI) \cite{EPI} was computed for both horizontal (HD) and vertical (VD) directions. As shown in Table~\ref{tab:epi}, the proposed method consistently achieves higher EPI values than SEGSID across all datasets, indicating improved preservation of meaningful structural boundaries while suppressing speckle noise.

\begin{table}[h]
\centering
\caption{EPI comparison between SEGSID and the proposed method. Higher values indicate better edge preservation.}
\label{tab:epi}

\setlength{\tabcolsep}{4pt}
\renewcommand{\arraystretch}{1.05}

\begin{tabular}{lcccc}
\hline
\multirow{2}{*}{Dataset} & \multicolumn{2}{c}{SEGSID} & \multicolumn{2}{c}{Proposed} \\
\cline{2-5}
 & HD & VD & HD & VD \\
\hline
KLSG   & 0.9093 & 0.9284 & \textbf{0.9338} & \textbf{0.9519} \\
DEBRIS & 0.9582 & 0.9605 & \textbf{0.9813} & \textbf{0.9783} \\
URPC   & 0.9268 & 0.9164 & \textbf{0.9377} & \textbf{0.9278} \\
\hline
\end{tabular}
\end{table}

\begin{figure}[htbp]
\centering
\setlength{\tabcolsep}{2pt}

\hspace{0.05\linewidth}
{\small \textbf{Noisy}}
\hspace{50pt}
{\small \textbf{SEGSID}}
\hspace{50pt}
{\small \textbf{Proposed}}\\[-1pt]

\vspace{2pt}

\noindent
\makebox[0pt][l]{\hspace{-12pt}\rotatebox{90}{\small DEBRIS}}
\begin{tabular}{ccc}
\includegraphics[width=0.31\linewidth]{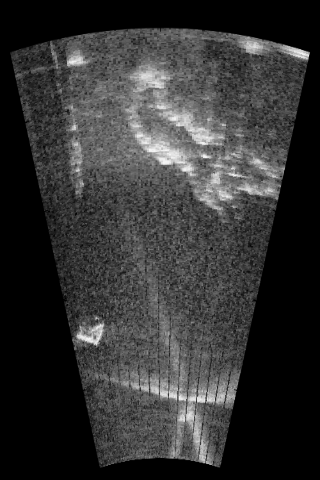} &
\includegraphics[width=0.31\linewidth]{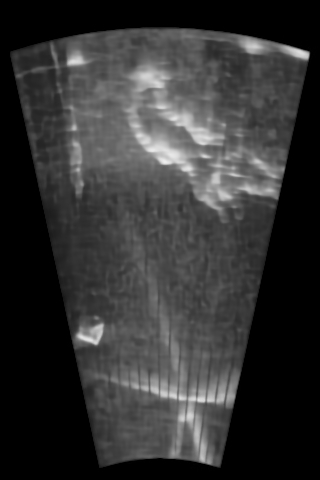} &
\includegraphics[width=0.31\linewidth]{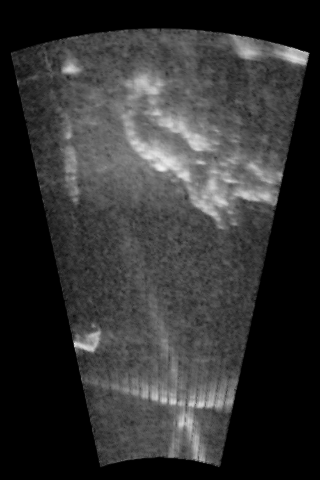} \\[-2pt]

{\scriptsize M: $\infty$ ,\ E:  1 /1} &
{\scriptsize M: 3.412 ,\ E:  0.95 /0.96} &
{\scriptsize M: 0.875 ,\ E:  0.98 /0.98}
\end{tabular}

\vspace{1pt}

\noindent
\makebox[0pt][l]{\hspace{-8pt}\rotatebox{90}{\small KLSG}}
\begin{tabular}{ccc}
\includegraphics[width=0.31\linewidth]{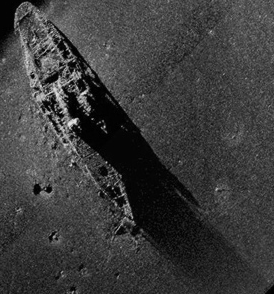} &
\includegraphics[width=0.31\linewidth]{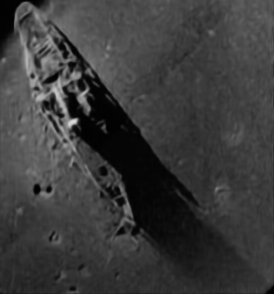} &
\includegraphics[width=0.31\linewidth]{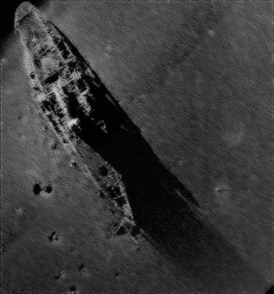} \\[-2pt]

{\scriptsize  M: $\infty$ ,\ E:  1 /1} &
{\scriptsize M: 3.685,\ E: 0.92 /0.93} &
{\scriptsize M: 2.0176,\ E: 0.94  /0.95}
\end{tabular}

\vspace{2pt}

\noindent
\makebox[0pt][l]{\hspace{-8pt}\rotatebox{90}{\small URPC}}
\begin{tabular}{ccc}
\includegraphics[width=0.31\linewidth]{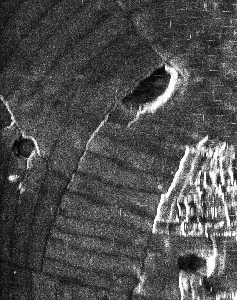} &
\includegraphics[width=0.31\linewidth]{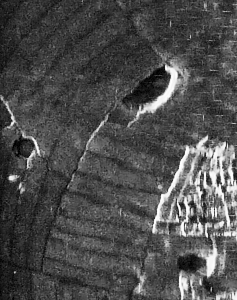} &
\includegraphics[width=0.31\linewidth]{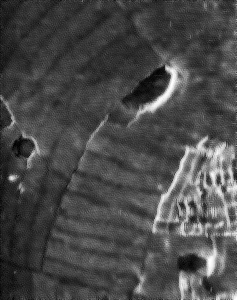} \\[-2pt]

{\scriptsize  M: $\infty$ ,\ E:  1 /1} &
{\scriptsize M: 2.595,\  E: 0.89 /0.90} &
{\scriptsize M: 1.368,\ E: 0.91 /0.91}
\end{tabular}

\vspace{1pt}

\noindent
\makebox[0pt][l]{\hspace{-12pt}\rotatebox{90}{\small SASSED}}
\begin{tabular}{ccc}
\includegraphics[width=0.31\linewidth]{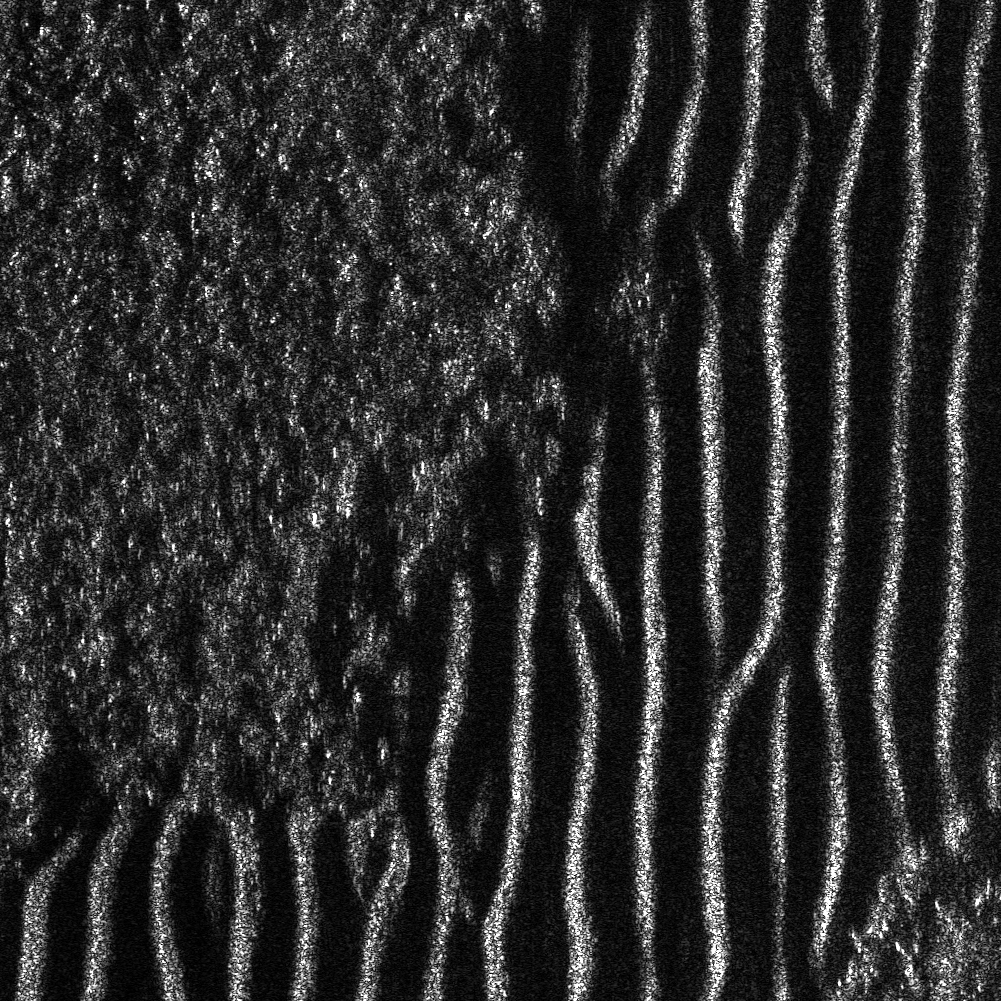} &
\includegraphics[width=0.31\linewidth]{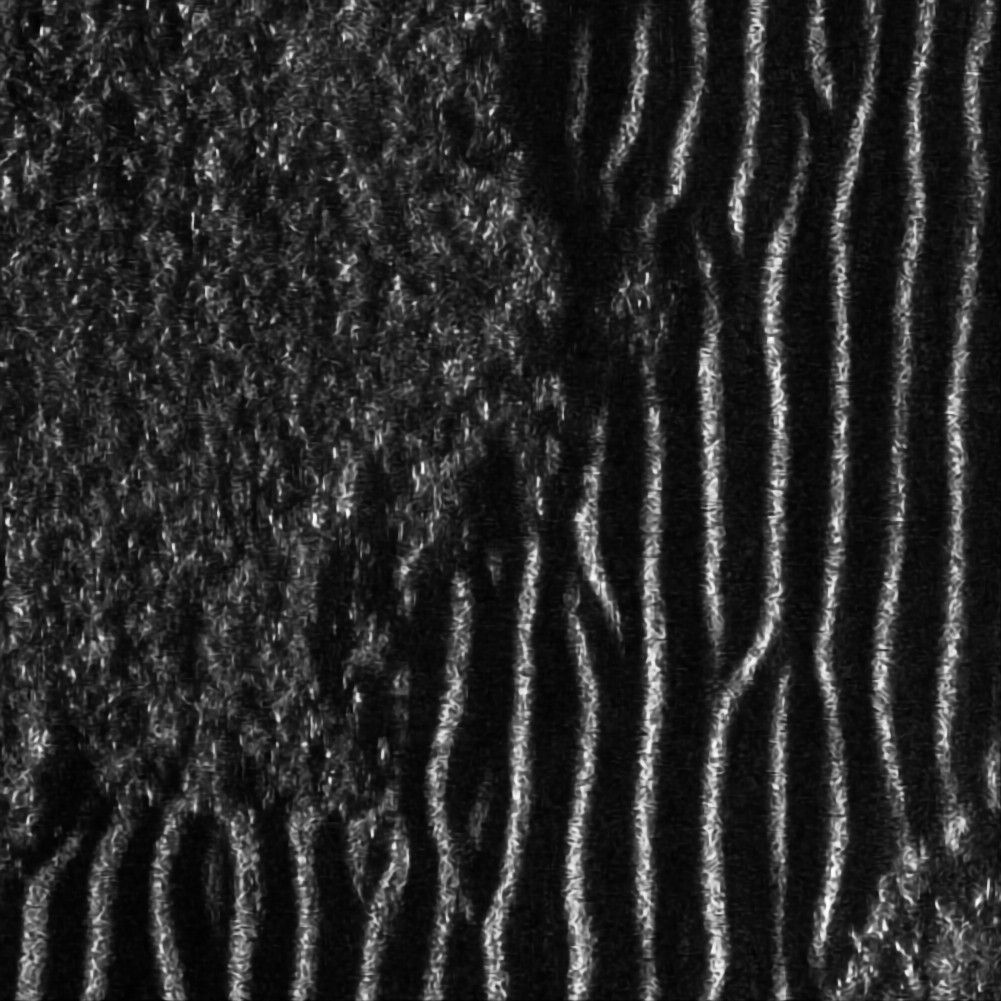} &
\includegraphics[width=0.31\linewidth]{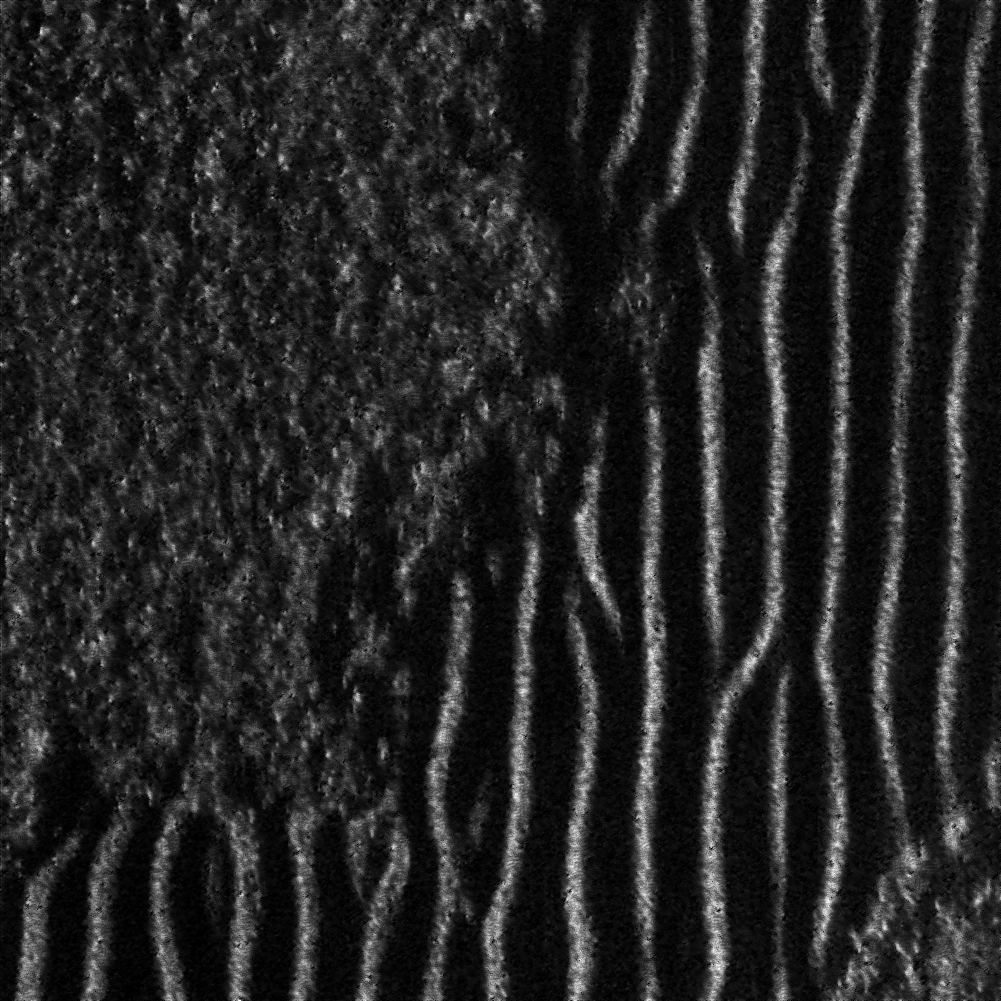} \\[-2pt]

{\scriptsize  M: $\infty$,\ E:  1 /1} &
{\scriptsize M:1.673,\ E: 0.74 /0.70} &
{\scriptsize M: 2.980,\ E: 0.84 /0.81}
\end{tabular}

\caption{Qualitative comparison across datasets. Columns show the noisy input, SEGSID result, and the proposed method (left to right). Values below denoised images denote M-score and EPI (HD/VD), respectively. The last two rows illustrate generalization results of the proposed method on unseen datasets.}
\label{fig:qual_minipage}
\end{figure}
The proposed method consistently achieves the lowest M-scores and high EPI across all evaluated datasets, outperforming the strongest existing baselines, including SEGSID and its knowledge-distilled variant.
 Notably, these gains are obtained using a compact model with only $0.16$\,M parameters, highlighting a favorable trade-off between performance and computational efficiency. To complement the quantitative evaluation, Fig.~\ref{fig:qual_minipage} illustrates qualitative results from representative samples across four sonar datasets. The proposed method achieves noticeable speckle reduction while maintaining structural details and object boundaries. Compared to existing approaches, the restored images exhibit improved visual clarity in homogeneous regions without compromising edge definition, indicating consistent behavior across datasets. Comprehensive ablation studies on the loss components and target variance, along with detailed implementation settings, are provided in the supplementary material.

\subsection{Cross-Dataset Evaluation on SAS Data}

To assess generalization, we perform a cross-dataset evaluation on the SASSED dataset, which consists of synthetic aperture sonar (SAS) imagery.
The proposed model is trained on the DEBRIS dataset and directly evaluated on SASSED without retraining or fine-tuning, with the target variance $\sigma_{\text{tgt}}^2$ fixed at $0.2838$. This setting introduces a pronounced domain shift due to the fully developed Gamma-distributed speckle characteristics inherent to SAS data. Table~\ref{tab:sassed_metrics} reports the M-score and EPI comparison on SASSED. Despite the substantial mismatch in speckle statistics and imaging modality, the proposed method outperforms SEGSID-KD (trained on DEBRIS), achieving significantly lower M-scores and higher EPI. These results demonstrate that the proposed statistical modeling framework captures intrinsic speckle characteristics and generalizes effectively across heterogeneous coherent imaging systems without domain-specific adaptation.

\begin{table}[h]
\centering
\caption{M-score and EPI comparison on the SASSED dataset.}
\label{tab:sassed_metrics}

\setlength{\tabcolsep}{6pt}
\renewcommand{\arraystretch}{1.1}

\begin{tabular}{lccc}
\hline
\textbf{Method} & \textbf{M-score} & \textbf{EPI-HD} & \textbf{EPI-VD} \\
\hline
SEGSID & 1.927 & 0.7488 & 0.7053 \\
\textbf{Proposed} & \textbf{1.178} & \textbf{0.8443} & \textbf{0.8119} \\
\hline
\end{tabular}

\end{table}
\subsection{Computational Efficiency}
Computational efficiency is evaluated using throughput (images/s), trainable parameters, and MACs on an NVIDIA RTX A4500 GPU with input resolution $160\times160$ (Table~\ref{tab:computation}). The proposed method is highly compact (0.160\,M parameters) and substantially lighter than SEGSID and SEGSID-KD. Although SEGSID-KD achieves the highest throughput through knowledge distillation, the proposed method attains comparable speed (532.76 images/s) without distillation or pruning, while requiring only 4.14\,G MACs. These results demonstrate an efficient trade-off between computational cost and despeckling performance, supporting real-time sonar deployment.

\begin{table}[h!]
\centering
\caption{Computational efficiency comparison at an input resolution of $160\times160$.}
\label{tab:computation}
\setlength{\tabcolsep}{6pt}
\renewcommand{\arraystretch}{1.1}
\begin{tabular}{lccc}
\toprule
\textbf{Method} & \textbf{\#Param (M)} & \textbf{MACs (G)} & \textbf{Throughput (images/sec)} \\
\midrule
SEGSID      & 23.0 & 1652.58 & 3.07 \\
SEGSID-KD   & 1.38 & \textbf{2.22}    & \textbf{633.05} \\
{Proposed} & \textbf{0.16} & 4.14 & 532.76 \\
\bottomrule
\end{tabular}
\end{table}


\section{Implementation Details}

The proposed framework is implemented in PyTorch~2.6.0 with CUDA~12.4 and trained on an NVIDIA RTX~A4500 GPU (24\,GB VRAM). Optimization is performed using AdamW with a fixed learning rate of $1\times10^{-5}$, a batch size of 8, and 50 training epochs. Training is conducted on randomly extracted $64\times64$ patches from the noisy DEBRIS and KLSG datasets. To ensure stable self-supervised convergence, a median-guided prior is employed during the initial 30 epochs, after which optimization is primarily driven by the proposed physics-informed statistical and structural constraints.

\subsection{Speckle Augmentation Strategy}

To improve robustness across varying noise levels, synthetic multiplicative speckle is applied to 50\% of training samples using a Gamma distribution $\Gamma(L, 1/L)$, where $L \in \{1,2,3,4\}$. This exposes the network to diverse speckle conditions while preserving the underlying reflectivity structure. As shown in Table~\ref{tab:speckle_aug}, the ablation study on the KLSG dataset demonstrates that speckle augmentation improves performance compared to training without augmentation.

\begin{table}[h]
\centering
\caption{Effect of speckle augmentation on the KLSG dataset.}
\label{tab:speckle_aug}

\setlength{\tabcolsep}{10pt}
\renewcommand{\arraystretch}{1.1}

\begin{tabular}{lc}
\hline
\textbf{Configuration} & \textbf{M-score} \\
\hline
Without Speckle Augmentation & 6.402 \\
With Speckle Augmentation & \textbf{6.142} \\
\hline
\end{tabular}

\end{table}
\section{Extended Ablation Studies}
\subsection{Effect of Target Variance}

\begin{figure}[h]
    \centering
    \includegraphics[width=\linewidth,height=0.35\linewidth,keepaspectratio]{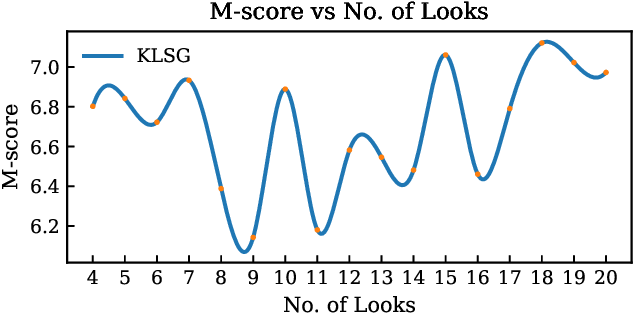}
    \caption{Ablation analysis of the target variance parameter showing M-score trends across KLSG dataset.}
    \label{fig:Target_Variance}
\end{figure}

An ablation study on the KLSG dataset evaluates the effect of the target variance parameter. Fig.~\ref{fig:Target_Variance} shows that M-score improves within a moderate target variance range, while larger values lead to gradual degradation due to the trade-off between speckle suppression and structural detail preservation.

\subsection{Structural Loss Weight Sensitivity}

\begin{figure}[h]
    \centering
    \includegraphics[width=\linewidth,height=0.34\linewidth,keepaspectratio]{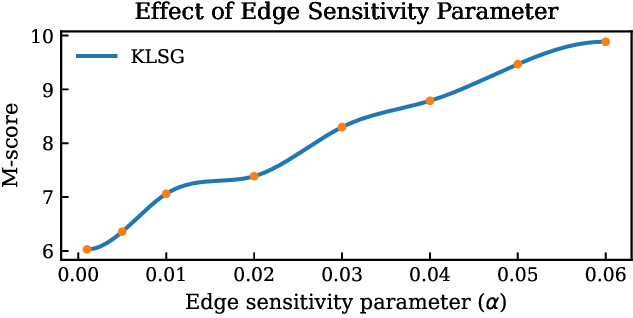}
    \caption{Ablation study on the structural loss weight showing M-score trends across KLSG dataset}
    \label{fig:structural_loss}
\end{figure}

The structural term uses weighted gradient consistency with edge-aware exponential weighting to preserve edges while suppressing speckle in homogeneous regions. Fig.~\ref{fig:structural_loss} shows that lower structural weights provide insufficient regularization, whereas larger weights cause over-smoothing and increased M-scores. Optimal performance is achieved at small weights, with a consistent trend observed across all datasets.

\subsection{Loss Ablation on KLSG}

The impact of different loss formulations is evaluated on the KLSG dataset with the target variance fixed to 0.1175. Table~\ref{tab:KLSG_ablation} reports the corresponding M-scores and EPI metrics. While the median-only baseline provides reasonable performance, incorporating statistical regularization yields a notable improvement. The combined statistical and structural regularization achieves the lowest M-score and the best edge preservation performance, demonstrating their complementary effect.

\begin{table}[h]
\centering
\caption{Loss ablation and edge preservation results on the KLSG dataset (Target Variance = 0.1175).}
\label{tab:KLSG_ablation}
\setlength{\tabcolsep}{5pt}
\renewcommand{\arraystretch}{1.1}
\begin{tabular}{lccc}
\hline
\multirow{2}{*}{Loss Formulation} & \multirow{2}{*}{M-score} & \multicolumn{2}{c}{EPI} \\
\cline{3-4}
 &  & HD & VD \\
\hline
Median only (Baseline) & 7.668 & 0.906 & 0.926 \\
Median + Statistical & 6.377 & 0.926 & 0.946 \\
Median + Structural & 7.449 & 0.908 & 0.928 \\
\begin{tabular}[c]{@{}l@{}}
Median + Statistical + \\
Structural (Full)
\end{tabular}
& \textbf{6.142} & \textbf{0.933} & \textbf{0.951} \\
\hline
\end{tabular}
\end{table}

\section{Conclusion}
\label{sec:conclusion}

This letter presented a physics-guided self-supervised framework for sonar image
despeckling in the homomorphic domain. Despeckling is formulated through residual
consistency under variance-targeted statistical constraints and edge-aware structural
regularization, using a lightweight network design. The proposed model achieves
state-of-the-art performance on real sonar and SAS datasets and demonstrates strong
cross-dataset generalization without retraining or clean supervision.

\section{Acknowledgments}
 We extend our sincere gratitude to the Indo-French Centre for the Promotion of Advanced Research (CEFIPRA) under the project IFC/SARI/2022/7143 and the Indian Institute of Technology Goa (IIT Goa) for their financial support.

\bibliographystyle{IEEEtran}
\bibliography{refs}

\end{document}